\documentclass[a4]{article}
\usepackage{graphicx,psfrag,epsf}
\usepackage{enumerate}
\usepackage[square,sort,comma,numbers]{natbib}
\bibpunct{(}{)}{;}{a}{}{,}
\usepackage{url}
\usepackage{amsmath, amsthm, amssymb,amsfonts}

  \newlength\titlebox  
    \clearpage
 \twocolumn 
\renewenvironment{abstract}{\centerline{\large\bf Abstract}\vspace{0.5ex}\begin{quote}}{\par\end{quote}\vskip 1ex} 
\DeclareMathOperator{\Poi}{Poi}
\DeclareMathOperator{\Unif}{Unif}
\DeclareMathOperator{\Exp}{Exp}

\DeclareMathOperator{\argmin}{argmin}
\DeclareMathOperator{\argmax}{argmax}

\DeclareMathOperator{\Tr}{Tr}
\DeclareMathOperator{\obs}{obs}
\DeclareMathOperator{\mis}{mis}

\def\glasso{{\rm glasso}}
\def\av{{\rm av}}
\def\pair{{\rm pair}}
\def\naive{{\rm naive}}
\def\loh{{\rm LW}}
\def\inv{^{-1}}
\def\uv{{(u,v]}}

\newcommand{\BR}{{\mathbb{R}}}

\newcommand{\CA}{{\cal A}}

\newcommand{\CO}{{\cal O}}

\begin{document}

%%%%%%%%%%%%%%%%%%%%%%%%%%%%%%%%%%%%%%%%%%%%%%%%%%%%%%%%%%%%%%%%%%%%%%%%%%%%%%

\title{\bf Change point detection for graphical models in the presence of missing values}
\author{Malte Londschien\footnote{The first two authors contributed equally to this work.}, \ 
Solt Kov\'acs\footnotemark[1], \
Peter B\"uhlmann \\
Seminar for Statistics, ETH Z\"urich, Switzerland}

\maketitle
\thispagestyle{empty}

\bigskip
\begin{abstract}
We propose estimation methods for change points in high-dimensional covariance structures with an emphasis on challenging scenarios with missing values. We advocate three imputation like methods and investigate their implications on common losses used for change point detection. We also discuss how model selection methods have to be adapted to the setting of incomplete data. The methods are compared in a simulation study and applied to a time series from an environmental monitoring system. An implementation of our proposals within the \textsf{R}-package \textbf{hdcd} is available via the Supplementary materials.
\end{abstract}


\section{Introduction}
The area of high-dimensional inverse covariance (or precision) matrix estimation has developed considerably over the past years. This is partly due to the off-diagonal entries of the precision matrix encoding un-scaled partial correlations and in the case of a multivariate normal distribution even conditional independence between two variables given all the others. The resulting graph, with variables as nodes and edges for nonzero off-diagonal entries is called a conditional independence graph or Gaussian graphical model (GGM), respectively \citep{lauritzen}.

In a lot of settings where the graphical structure of a set of observations is of interest, the data is in the form of a time series (or has another natural ordering, e.g. by space or genome location) and the underlying distribution might change over time. A lot of attention is given to the setting where there are structural breaks, so called change points, in between which the observations are identically distributed. Disregarding this structure and estimating a GGM for all observations leads to a fit of mixtures that does not model the true GGM well for any time point. In such cases to estimate the graphical structure, prior good estimation of the change points is key. 

Real world data sets often suffer from missing values. In the presence of change points and without prior knowledge of them, usual imputation methods are not expected to perform well as they require observations from a homogeneous distribution. However, some kind of imputation method is necessary in order to apply any of the current methods to find change points, between which the distribution can be assumed to be homogeneous, leading to a chicken and egg kind of problem. In practice, we observe that the approach of naively imputing mean values columnwise (i.e.~for each variable separately) on the whole data set and then estimating change points does not perform well. For example, if observations are not missing completely at random, but according to some structure (e.g.~in some blockwise manner due to the simultaneous failure and repair of multiple sensors), this naive approach tends to (falsely) detect change points corresponding to the missingness pattern rather than the true changes in the underlying signal (see Section \ref{section:simulations} for more description and simulation results regarding this naive approach). We will demonstrate that integrating imputation and change point detection in a unified framework can lead to substantial gains.

\subsection{Related work}
For homogeneous observations, \citet{MeinsBuhl} proposed nodewise regression using the Lasso \citep{lasso_original} to recover the conditional independence structure from data. \citet{YuanLin} introduced an estimator for the precision matrix (and thus the GGM) via maximisation of the $L_1$-penalized Gaussian log-likelihood over the set of positive definite matrices. The graphical Lasso (glasso) algorithm of \citet{glasso} with subsequent improvements (\citealp{witten_glasso}; \citealp{mazumder2012}) gained popularity for computing such estimates and the term glasso is now also associated with the corresponding estimator. Computational approaches for the estimator were also presented by \citet{Banerjee_dAspremont}. Theoretical properties were investigated by \citet{YuanLin}, \citet{Rothman}, \citet{sparsistency} and \citet{Ravikumar}.

For the non-homogeneous case, \citet{ZhouLaffertyWassermann} considered cases where the graph structure varies smoothly and \citet{KolarXing} investigated under which conditions the graph structure can be recovered consistently. In a setting with abrupt structural breaks, \citet{Kolar_Gauss_estim} proposed to estimate the locations of change points with a total variation penalty for consecutive observations either using nodewise regression or a penalized likelihood approach. \citet{GibberdNelson} proposed the group-fused graphical Lasso (with a Frobenius norm fusion penalty) and compared it to $L_1$-fused methods that they called independent-fused graphical Lasso. Some statistical analysis of the group-fused graphical Lasso was provided by \citet{Gibberd_Roy}. A penalized likelihood approach was proposed by \cite{Angelosante}, seeking the corresponding optimal partitioning via dynamic programming.
While \citet{Kolar_Gauss_estim} used an accelerated gradient descent method, \citet{GibberdNelson} proposed
an ADMM algorithm to calculate the estimator, similarly to
\citet{Hallac_network}, who considered larger classes of fusion type penalties.
Other computational approaches  include an approximate majorize-minimize 
algorithm \citep{Bybee_Atchade_JMLR} or a specific 
greedy search \citep{Hallac_GGS}.
More thorough statistical analysis of similar proposals was conducted by \citet{RoyMichailidis}, \citet{Avanesov_theory}, \citet{Dette_theory} and \citet{Wang_Yu_Rinaldo_theory}. 

In the presence of missing values, the maximisation of the $L_1$-penalized log-likelihood (see equation (\ref{eq:loglikelihood_missing_1}) later on) is no longer a convex problem, even for homogeneous observations. For this scenario of homogeneous observations with missing values, \citet{missglasso} proposed to use an Expectation Maximisation algorithm coupled with the glasso to obtain an estimate of the precision matrix. The single example of the treatment of missing values in the context of change point detection we are aware was done by \citet{changepoint_missing}.
They however considered an online (sequential) setup and assumed that the observations lie close to a time-varying low-dimensional submanifold within the observation space. This assumption is appropriate in e.g. video surveillance, but it is unrealistic in the setting of GGMs, even if the underlying precision matrix is assumed to be sparse. In this latter setting of GGMs there is currently no available method for change point detection incorporating missing values. We aim to fill this gap.

\subsection{Possible applications}
Applications of change point detection within graphical models include the analysis of environmental measurements, biological data and financial time series, which potentially encounter the problem of missing values. 
A specific motivating example for our proposals was the shallow groundwater monitoring data set, which we will discuss later on. To mention further concrete examples, multivariate change point detection methods (in GGMs) could be useful for example in detecting the signal of abrupt climate changes imprinted simultaneously in multiple climate proxy records, e.g. ice cores in Antarctica or speleothems, all facing missing values \citep[see e.g.,][]{s1}.

So far, it is common practice to discard variables with too much missingness and to perform simple imputations on the rest \citep[see e.g.,][]{ecp}. Also, sometimes univariate methods are used to detect change points for each variable separately \citep[see e.g.,][]{s2}. These approaches are suboptimal. Discarding non-complete observations, especially with high-dimensional data and inhomogeneity of the missingness structure with respect to time, is impractical as it results in a significant loss in information, possibly leaving only a handful of observations. Discarding all variables with missing observations can lead to keeping only a fraction of variables. Both approaches might lead to no observations for an analysis in an extreme case.

\subsection{Our contribution}
Our investigated setup is composed of three problems: change point detection, estimation of GGMs and the treatment of missing values. While there are many possible applications, there 
is currently no readily available method combining the three and thus capable of estimating change points in (high-dimensional) GGMs in the presence of missing values. We fill this gap and provide practitioners with practically usable, and in particular computationally tractable  methods that work both in the presence and in the absence of missing values. We provide an implementation of these methods within the \textsf{R}-package \textbf{hdcd}, see the Supplementary materials.

We investigate different scenarios of missingness (both missing completely at random and with structures resembling real world scenarios), discuss the resulting difficulties and propose viable estimation approaches. Their performance is evaluated in a simulation study and applied to data from an environmental monitoring system.

\section{Change point detection without missing values}
\label{section:complete_case}
Consider a sequence of independent Gaussian random variables $(X_i)_{i = 1}^n \in \BR^p$ with means $\mu_i$ and covariance matrices $\Sigma_i = \Omega_i\inv$ such that the map $i \mapsto (\mu_i, \Sigma_i)$ is piecewise constant. 
Let 
\begin{equation*}
\alpha^0 := \{0, n\} \cup \{i : (\mu_i, \Sigma_i) \neq (\mu_{i + 1}, \Sigma_{i + 1})\}
\end{equation*}
be the set of segment boundaries.
We label the elements of $\alpha^0$ by their natural order starting with zero such that consecutive pairs of elements in $\alpha^0$ define  segments $(\alpha^0_{k-1}, \alpha^0_{k}],\ k = 1, \dotsc, |\alpha^0| - 1$ within which the $X_i$ are i.i.d. For $0 \leq u < v \leq n$ let $X_{(u,v]}$ denote the matrix of the observations
$X_{u+1},\ldots,X_{v}$, denote with $\hat\mu^{(u,v]}$ their mean and let $S_{(u,v]} := (X_{(u,v]} - \hat\mu^{(u,v]})^T (X_{(u,v]} - \hat\mu^{(u,v]})/(v - u)$ be the corresponding covariance matrix.

For $\delta > 0$ define $\CA_{n,\delta}$ as the family of possible sets of segment boundaries such that the minimal segment length is not smaller than $\delta n$.
Let $L_n((u,v])$ be some loss after fitting an adequate model to $X_{(u,v]}$ that is normalized by $n$ and scales with the segment length $v - u$, see e.g.~Equation~\eqref{eq:glasso_estimator_definition}. For a penalty parameter $\gamma > 0$ for the number of segments, an estimator for $\alpha^0$~is
\begin{equation}
\label{eq:changepointestimator}
\hat\alpha^0 := \underset{\alpha \in \CA_{n, \delta}}{\argmin} \sum_{j = 1}^{|\alpha| - 1} L_n((\alpha_{j-1}, \alpha_j]) + \gamma.
\end{equation}
This estimator can be computed using dynamic programming with $\CO(n^2)$ evaluations of $L_n$, and considerably faster when certain pruning steps are applicable (typically in scenarios with many change points, see \citealp{Killick_etal}). In particular without pruning steps (e.g.~in scenarios with a handful of change points only), this is computationally infeasible if $n$ is large, especially if the cost to evaluate $L_n$ is significant. 

Binary Segmentation \citep{Vostrikova} is a much faster greedy algorithm to estimate $\alpha^0$. For this define the gains function of some segment $(u,v]$ at some split point $s$ to be 
\begin{equation}
G_n^{(u,v]}(s) := L_n((u, v]) - \left(L_n((u, s]) + L_n((s, v])\right)
\label{eq:gains_fun}
\end{equation}
and define
\begin{equation}
    \hat\alpha_{(u,v]} := \underset{s \in \{u + \delta n, \dotsc, v - \delta n\}}{\argmax} G_n^{(u,v]}(s).
    \label{eq:singlesplitpoint}
\end{equation}
The search for a single change point in $(X_i)_{i = 1}^{n}$ by solving (\ref{eq:changepointestimator}) breaks down to finding $\hat\alpha_{(0, n]}$ and checking if $G_n^{(0,n]}(\hat\alpha_{(0, n]}) > \gamma$. For the multiple change point case Binary Segmentation (BS) finds an approximate solution to (\ref{eq:changepointestimator}) by recursively splitting segments using (\ref{eq:singlesplitpoint}) 
until the resulting segment length is smaller than $2 \delta n$ such that a split is no longer allowed or the corresponding gain is not bigger than $\gamma$, the minimally required gain to split. BS typically requires asymptotically $\CO(n \log(n))$ evaluations of $L_n$. Due to its greedy nature, BS is not optimal in terms of statistical estimation (as opposed to computing the optimal partitioning from \eqref{eq:changepointestimator} using dynamic programming). Wild Binary Segmentation (WBS,  \citealp{Fryzlewicz_WBS}) as well as the recently proposed Seeded Binary Segmentation (SeedBS) method of \cite{SeedBS} improve on statistical detection by evaluating the gains function \eqref{eq:gains_fun} for various random (for WBS) or deterministically constructed (for SeedBS) background intervals, each leading to a candidate split point via \eqref{eq:singlesplitpoint}. Out of the list of candidates the final set of change point estimates is subsequently derived. The improved detection is due to the fact that some of the generated background intervals only contain a single change point and in that case the detection is easier. While WBS loses some of the computational efficiency of plain BS (e.g.~in scenarios with many change points or very short spacing between some change points when one needs to draw a large number of random intervals), the more efficient deterministic interval construction of SeedBS requires only $\CO(n \log(n))$ evaluations of $L_n$ independent of the number of change points and thus SeedBS is similarly fast as plain BS.

Nonetheless, $\CO(n \log(n))$ evaluations of $L_n$ for BS can still be prohibitive if the cost of evaluating $L_n$ is large (e.g.~for high-dimensional model fits) and $n$ is big. Instead of evaluating $G_n^{(u,v]}(s)$ at every possible split in a full grid search to find its maximum, one can find (up to very few exceptions with special signal cancellation effects) one of its local maxima with adaptively chosen $\log(n)$ evaluations using the Optimistic Search strategies of \citet{OBS}. The expected gain curve for common losses (e.g. the squared error loss) is piecewise convex between the true underlying segment boundaries, such that in particular all local maxima correspond to change points. Hence, splitting at a local maximum instead of the global maximum, one does not induce a false discovery and the missed global maximum can still be found in a later step. Doing BS with this Optimistic Search is called Optimistic Binary Segmentation (OBS, \citealp{OBS}) and approximately requires $\CO(|\alpha^0|\log(n))$ evaluations of~$L_n$. Note that we only use the so-called naive variant of Optimistic Search and thus the naive variant of OBS in this paper (e.g.~simulation results). In practice, the observed gain curve (resulting from a single draw) is a noisy version of the expected one. As a consequence, the idealized piecewise convex structure of the expected gain curve is distorted and only conserved approximately (in terms of its rough shape). In very noisy scenarios the Optimistic Search can get stuck in a noise induced local maximum that is far away from one corresponding to a change point. Optimism is thus needed in noisy scenarios. In contrast to Optimistic Search (used in OBS), the full grid search (used in BS) finds the global maximum of the noisy observed gain curve, which is typically closer to one of the true underlying change points (provided that there is a sufficiently large jump in the underlying signal). We will see later on how the noisiness of the observed gain curve influences the applicability of OBS compared to BS in scenarios with lots of missing data.

Since the writing of this paper, \citet{OBS} proposed two new variants, the advanced and combined Optimistic Search. These have improved performance and stronger theoretical results for challenging scenarios with change points close to the segment boundaries. \citet{OBS} also describe Optimistic Seeded Binary Segmentation (OSeedBS), a combination of Seeded Binary Segmentation and Optimistic Search. OSeedBS is typically somewhat more costly than OBS, but as advantages, OSeedBS would be easy to parallelize and also has stronger statistical guarantees than BS (and thus also OBS). These more recent search methods could be combined with our later described methodology for detecting change points for GGMs and to increase flexibility, we added some of them to our \textsf{R}-package \textbf{hdcd}.

\citet{LeonBuhl} applied BS to a high-dimensional linear regression change point problem using the negative log-likelihood for Gaussian errors (sum of squared errors) resulting from a Lasso fit as a loss measure. They provided (under technical conditions) consistency results if the penalization parameter $\lambda$ for the Lasso is adjusted by the inverse of the square root of the relative segment length, i.e. by using $\sqrt{n / (v- u)}\lambda_0$ for some fixed $\lambda_0$ for the segment $(u,v]$.

In order to adapt this approach to the multivariate normal case, we set
\begin{equation}\label{eq:glasso_estimator_definition}
    L_n(\Omega; S_{(u,v]}) = \frac{v-u}{n}\left(\Tr(\Omega^T S_{(u,v]}) - \log(|\Omega|) \right).
\end{equation}
This way $L_n(\Omega; S_{(u,v]})$ is (up to a constant) the negative log-likelihood of a Gaussian with precision matrix $\Omega$ given observations $X_\uv$ that is scaled with segment length. As $L_n$ scales with $(v-u)/n$, we set $\lambda_{(u,v]} := \sqrt{(v-u)/n} \lambda_0$ and define the estimator
\begin{equation}
\label{eq:glasso_estimator}
\begin{split}
    \hat\Omega_{(u,v]}^\glasso 
&:= \underset{\BR^{p \times p} \ni \Omega \succ 0}{\argmin} L_n(\Omega; S_{(u,v]}) + \lambda_{(u,v]} \|\Omega\|_1\\
&= \underset{\BR^{p \times p} \ni \Omega \succ 0}{\argmin} \Tr(\Omega^T S_{(u,v]}) - \log(|\Omega|) \\
&\hspace{2.5cm}+ \sqrt{n/(v-u)}\lambda_0 \|\Omega\|_1.
\end{split}
\end{equation}
Hence, $\hat\Omega_{(u,v]}^\glasso$ is the glasso estimator from \citet{glasso} for the precision matrix of observations $X_\uv$ with penalization parameter $\sqrt{n/(v-u)}\lambda_0$. We thus replicate the scaling approach by \citet{LeonBuhl}, but use a different notation more suited to our GGM setup. We prefer not to penalize the diagonal in $\|\Omega\|_1$ and then use the in sample loss
\begin{equation*}
    L_n((u,v]) := L_n(\hat\Omega_{(u, v]}^\glasso; S_{(u,v]})
\end{equation*}
after fitting the glasso in equation (\ref{eq:gains_fun}) as the loss when greedily searching for optimal splits with BS or OBS.

We choose the graphical Lasso for the estimation of the precision matrix since some sparsity assumption seems crucial in high-dimensional scenarios. Note that \citet{Avanesov_theory} also relied on the graphical Lasso (or similar procedures for sparse precision matrices) and \citet{RoyMichailidis} also had a sparsity assumption in their change point detection proposals. Additionally, the change point detection is expected to be more powerful and reasonable when placing the sparsity assumption on the precision matrix rather than the covariance matrix, in particular when considering sparse changes between conditional dependencies among a few variables. For low-dimensional scenarios non-sparse (e.g. ridge type) estimators could also be reasonable, but differences in estimated change point locations compared to our glasso based approach are expected to be small.

Overall the algorithms have three tuning parameters: $\delta$ for the minimal relative segment length, $\gamma$ to control for the number of segments and $\lambda_0$ for the sparsity of the precision matrices. 
Note that the procedure is similar to the one proposed by \citet{Angelosante}, but as a crucial difference their penalty is not scaled according to the segment length with $\lambda_{(u,v]}$, and they include the penalty term into the gains function, which seems unnatural.

\section{Adapting to missing data}
\label{sec:adapting_to_missing_data}
For non-complete data denote with $x_{\obs, i}$ the observed part of the $i$-th observation and with $\Omega_{\obs,i}$ and $\mu_{\obs,i}$ the submatrix of $\Omega$ and the subvector of $\mu$ corresponding to the observed variables of $x_i$. Similarly denote with $\mu_{\mis, i}$ the subvector of $\mu$ corresponding to the variables of $x_i$ with missing values. Note that the observed components can differ across observations $x_i$. The normalized negative Gaussian log-likelihood of observations from a segment $(u, v]$ for a given mean $\mu$ and precision matrix $\Omega$ is
\begin{equation}
\label{eq:loglikelihood_missing_1}
\begin{split}
\ell (&
    \mu, \Omega, (x_i)_{i = u + 1}^v) 
= \frac{1}{2n}\sum_{i = u + 1}^v \Big(- \log(|2\pi\Omega_{\obs,i}|) \\
&+ (x_{\obs, i} - \mu_{\obs,i})^T \Omega_{\obs,i} (x_{\obs, i} - \mu_{\obs,i}) \Big) ,
    \end{split}
\end{equation}
treating observations with missing values as if they arose from a lower-dimensional Gaussian (by considering only the log-likelihood corresponding to the observed parts, i.e. submatrices of $\Omega$ and subvectors of~$\mu$). Set $\hat\mu^{(u,v]}$ to be the empirical mean of the observed part of $X_{(u,v]}$ (discarding the missing values). Note that $\hat\mu^\uv$ might have missing values itself if for some variable there is no observed value in the segment $\uv$. In order to simplify notation in terms of the mean vector, by plugging $\hat\mu^{(u,v]}$ into \eqref{eq:loglikelihood_missing_1}, let 
\begin{equation}
\ell(\Omega; (x_i)_{i = u + 1}^v) := \ell(\hat\mu^{(u,v]}, \Omega;  (x_i)_{i = u + 1}^v),
\label{eq:loglikelihood_missing_2}
\end{equation}
still evaluating the log-likelihood according to the observed parts. The $L_1$-penalized maximum likelihood estimator for observations $X_\uv$ is then
\begin{equation*}
\hat\Omega_{(u,v]} = \underset{\BR^{d\times d} \ni \Omega \succ 0}{\argmin}
\ell(\Omega; (x_i)_{i = u + 1}^v) + \frac{\lambda_{(u,v]}}{v-u}\sum_{u+1}^v \|\Omega_{\obs,i}\|_1.
\end{equation*}
Contrary to (\ref{eq:glasso_estimator}), this problem is no longer convex and cannot be solved efficiently with an update-based approach like the glasso. The Miss-Glasso algorithm proposed by \citet{missglasso} combines the glasso and an Expectation Maximization (EM) algorithm to estimate the precision matrix
in the presence of missing data. However, the algorithm needs complete observations for a good initialization, is computationally expensive due to a new glasso fit for each EM iteration and might get stuck in local optima. These features are especially critical in our setting of high-dimensional change point detection. High computational cost is prohibitive since we do a lot of evaluations of the loss function. More importantly even, for each split, Miss-Glasso would be initialised slightly differently and in some situations it could converge to a very different local optimum. This would result in jumps in the gain curve from equation (\ref{eq:gains_fun}) for some neighboring splits $s$.

While an accurate estimate $\hat\Omega_{(u,v]}$ (as aimed e.g. by Miss-Glasso) is necessary for a good fit, this is not needed for change point detection. It is sufficient to roughly approximate the piecewise convex shape of the (expected) gains function (\ref{eq:gains_fun}), as this makes sure that pronounced local maxima lie in a neighborhood of the true underlying change points. This key idea helps to avoid the heavy computations and local optima of the EM algorithm when doing BS or~OBS.
In the following, we propose estimators $\tilde S_{(u,v]}$ that approximate $S_{(u,v]}$ based on $(x_{\obs, i})_{i = u+1}^v$. We can then use $\tilde S_{(u,v]}$ in the glasso estimator (\ref{eq:glasso_estimator}) to obtain $\tilde\Omega_{(u,v]}^\glasso$ instead of $\hat\Omega^\glasso_{(u,v]}$ and use the resulting log-likelihood $\ell(\tilde\Omega^\glasso_{(u,v]}, (x_i)_{i = u+1}^v$) as in  equation (\ref{eq:loglikelihood_missing_2}) with minor modifications (see Section~\ref{section:curve_smoothing}) as loss measure.

\subsection{Missing value imputations}
\label{section:imputation_methods}
We propose three different estimators $\tilde S_{(u,v]}$ for data with missing values. These will lead to different results, computational costs and applicabilities.

\paragraph{The average imputation estimator}
In a first attempt, we impute the missing values $(x_{\mis, i})_{i = u+1} ^ v$ with the average value of the corresponding variables within the interval $(u,v]$. Thus define centered variables
\begin{equation*}
    z_{\obs, i}^\uv := x_{\obs,i} - \hat\mu^\uv_{\obs,i} \ 
    \textrm{ and } \ 
    \hat{z}^{(u,v]}_i := (\hat z^\uv_{\obs,i}, 0).
\end{equation*}
The latter notation indicates that the centered vector $\hat{z}^{(u,v]}$ is padded with zeros at the appropriate positions of missing entries. Then set 
\begin{equation*}
    {\tilde S}_{(u,v]}^\av := \frac{1}{v - u} (\hat z^\uv_{u+1}, \dotsc, \hat z^\uv_{v})^T (\hat z^\uv_{u+1}, \dotsc, \hat z^\uv_{v})
\end{equation*}
as the average imputation estimator of the in-segment covariance matrix. This underestimates the variance and covariance of variables where values are missing. Nonetheless the average method serves as a baseline.

\paragraph{The Loh-Wainwright bias corrected estimator}
We try to counteract the shortcomings of the average imputation method using the bias correction presented by \citet{loh2012structure}. The authors show that if the $j$-th variable of an observation in $X_{(u,v]}$ is discarded with probability $\rho_j$, setting
\[M_{i,j} : = 
\begin{cases}
\frac{1}{(1 - \rho_i) (1 - \rho_j)} & i\neq j \\
\frac{1}{1 - \rho_i} & i = j
\end{cases},
\]
the matrix $\tilde S_{(u,v]}^\av \circ M$ is an unbiased estimator of $S_{(u,v]}$. Here $\circ$ denotes the Hadamard (pointwise) product for matrices. In practice, we have to estimate the $\rho_i$ based on the proportion of missing observations, leading to $\hat M$. Note that the resulting matrix $\tilde S_{(u,v]}^\av \circ \hat M$ is not necessary positive semi-definite which is required for the glasso algorithm \citep{glasso} to converge. We thus compute the closest positive semi-definite matrix to $\tilde S_{(u,v]}^\av \circ \hat M$ with respect to the Frobenius norm using the Higham algorithm \citep{higham2002computing} and take this as our final Loh-Wainwright (LW) bias corrected estimator $\tilde S_{(u,v]}^\loh$. 
Note that the Higham algorithm has an asymptotic complexity of $\CO(p ^ 3)$ similar to the glasso and thus evaluating the gain using the LW estimator is computationally more expensive than the average imputation method.

Alternatively, one could plug in $\tilde S_{(u,v]}^\av \circ \hat M$ as the sample covariance matrix into equation (\ref{eq:glasso_estimator}), similar to what has been suggested by \cite{Kolar_missing} as an alternative to the Miss-Glasso procedure of \cite{missglasso} in scenarios with homogeneous distributions with missing data. While one can skip the calculation of the nearest positive definite matrix this way, the problem is that not all algorithms that have been developed to solve estimation problems from equation (\ref{eq:glasso_estimator}) work with input matrices that are not positive semi-definite. In particular, as mentioned previously, the glasso algorithm itself would not work either. As we intend to rely on standard algorithms and corresponding software that are fast (as several repeated fits are necessary for change point detection), we do not further pursue this alternative.

\paragraph{Pairwise covariance estimation}
A third estimate can be based on pairwise covariance estimates, where the covariance between two variables is calculated from observations where both corresponding variables are available. Similar to the LW estimator, this gives a valid estimate of the variances and covariances of variables, even if the missingness structure in the data is not homogeneous. If there are less than two complete observations for a pair of variables in a segment, we set the covariance between these variables to be zero. Again, as with the LW approach, this might yield a matrix that is not positive semi-definite. We hence apply the Higham algorithm to compute the closest positive semi-definite matrix and denote the resulting estimator by $\tilde S_\uv^\pair$.

\subsection{Avoiding jumps in the gain curve}
\label{section:curve_smoothing}
\begin{figure}[h]
    \centering
    \includegraphics[width = 230pt]{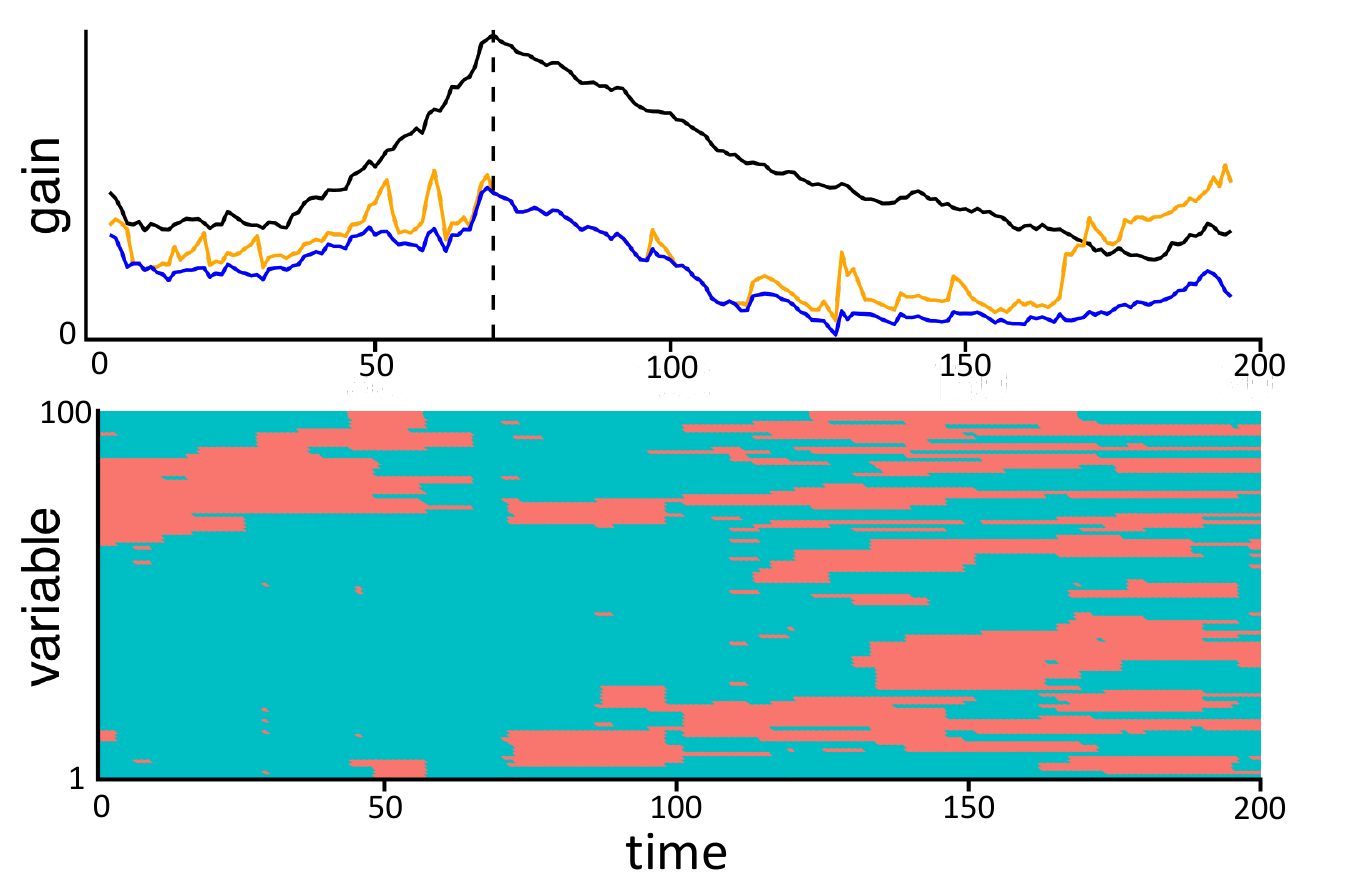}
    \caption{A scenario of simulated blockwise missing data (bottom). Red parts correspond to missing (i.e. deleted) values in the data matrix. The top shows the recovered gains if the full data without missing values is used (black), as well as in the case of missing values our proposed estimator using the LW method (blue), the naive version (orange) exhibiting jumps and the location of the true underlying change point (vertical dashed line).}
    \label{fig:curve_smoothing}
\end{figure}

For all of the three proposals above, the estimation of the variance of a variable on a given segment requires at least two observations with non missing values.
In order to obtain a meaningful estimate,  more non-missing observations are necessary. 
 As a consequence, with a lot of missing values or small segments, we might only be able to estimate a submatrix of the full covariance matrix. For such segments, the log-likelihood can then only be computed for a submodel of the entire multivariate Gaussian distribution. When evaluating the gains function (\ref{eq:gains_fun}) at some split point $s$, it might thus happen that the log-likelihood of the segment $\uv$ is calculated based on a larger covariance matrix (and thus a model with more parameters) than for $(u, s]$ or $(s, v]$. As the log-likelihoods of multivariate Gaussians of different dimensions are not easily comparable, this is especially problematic for split points $s$ such that the estimated covariance matrix for the neighboring split $s + 1$ has a different size. In such scenarios, the gains curve often has jumps between $s$ and $s+1$ (see Figure \ref{fig:curve_smoothing}).
 To alleviate this, we propose to restrict the log-likelihood  of $\uv$ to the dimensions available for $(u, s]$ and $(s, v]$.
 To make this more concrete, let us introduce some notation.

Denote for a segment $\uv$ and some $k \geq 2$ with $j_\uv(k)$ the indices of the variables for which at least $k$ values are observed in the segment $\uv$. For any imputation method $* \in \{\av, \loh, \pair\}$ let ${\tilde S}_{(u,v]}^{*, k} : = ({\tilde S}_{(u,v]}^{*})_{j_\uv(k), j_\uv(k)}$ be the submatrix of ${\tilde S}_{(u,v]}^{*}$ where each variable has at least $k$ observed values. Finally denote with $\tilde\Omega^\glasso_{(u,v], k} = \tilde\Omega^\glasso_{(u,v]}({\tilde S}_{(u,v]}^{*, k})$ the obtained glasso fit for the submatrix. In our simulations we set the minimal number of required observations for keeping a variable to be $k = 5$.

\begin{figure}[ht]
    \centering
    \includegraphics[width = 230pt]{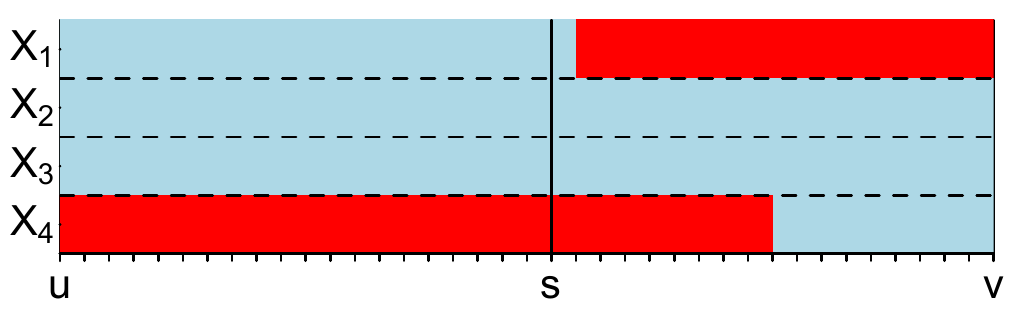}
    \vspace{-0.2cm}
    \caption{Illustrative example of a segment with $38$ observations of dimension $4$. Red blocks correspond to missing values and the black vertical line to a possible split.
    }
    \label{fig:blocks_plot}
\end{figure}

A naive estimator for the gains function is
\begin{equation*}
\begin{split}
    \tilde G^{(u,v], k}_{n, \naive}(s) :=\  
    & \ell(\tilde\Omega^\glasso_{(u,v], k}, (x_{i, j_{(u,v]}(k)})_{i = u + 1}^v) \ - \\
    & \ell(\tilde\Omega^\glasso_{(u,s], k}, (x_{i, j_{(u,s]}(k)})_{i = u + 1}^s) \ - \\
    & \ell(\tilde\Omega^\glasso_{(s, v], k}, (x_{i, j_{(s,v]}(k)})_{i = s + 1}^v).
    \end{split}
\end{equation*}
This naive estimator might be comparing log-likelihoods of multivariate normal distributions of different dimensions at some possible  split points $s$ if $j_{(u, s]}(k) \subsetneq j_\uv(k)$ or $j_{(s, v]}(k) \subsetneq j_\uv(k)$. 
See Figure~\ref{fig:blocks_plot} for an illustrative example. Here
$j_{(u, s]}(2) = \{1, 2, 3\}$, $j_{(s, v]}(2) = \{2, 3, 4\}$ and $j_\uv(2) = \{1, 2, 3, 4\}$.
The log-likelihood 
$\ell(\tilde\Omega^\glasso_{(u,v], 2}, (x_{i, j_{(u,v]}(2)})_{i = u + 1}^v)$
of the full segment would be using the first variable of the (s+1)-st observation for evaluation, whereas the log-likelihood  $\ell(\tilde\Omega^\glasso_{(s, v], 2}, (x_{i, j_{(s,v]}(2)})_{i = s + 1}^v)$ of the segment on the right would not.

To avoid this, we propose to use the slightly different estimator 
\begin{equation*}
\label{eq:adjusted_gain}
\begin{split}
    \tilde G^{(u,v], k}_n(s) :=\  
    & \ell(\tilde\Omega^\glasso_{(u,v], k}, (x_{i, j_{(u,s]}(k)})_{i = u + 1}^s) \ + \\
    & \ell(\tilde\Omega^\glasso_{(u,v], k}, (x_{i, j_{(s,v]}(k)})_{i = s + 1}^v) \ - \\
    & \ell(\tilde\Omega^\glasso_{(u,s], k}, (x_{i, j_{(u,s]}(k)})_{i = u + 1}^s) \ - \\
    & \ell(\tilde\Omega^\glasso_{(s, v], k}, (x_{i, j_{(s,v]}(k)})_{i = s + 1}^v).
    \end{split}
\end{equation*}
Here we only use the variables $j_{(u,s]}(k)$ and $j_{(s, v]}(k)$ in the calculation for the loss of the full segment $\uv$ when splitting at $s$. Hence, slightly different losses are used for the full segment depending on the split point. As we are not primarily interested in a segments' own loss, but rather a fair estimate of the gain, this is a sensible approach.

This does not increase the computational cost significantly, as for repeated evaluations of $\tilde G^{(u,v],k}_n$ we can keep the initial estimate $\tilde\Omega^\glasso_{(u,v]}$. Without this mechanism the gain curve exhibits jumps at the boundaries of missing blocks, as illustrated in Figure \ref{fig:curve_smoothing}. This might prevent BS and OBS from correctly estimating the change points. The severity of this problem depends on the block size. When the missingness structure is homogeneous in time, the block size is small such that this is not a pronounced issue. However, in many applications one faces blockwise missing data, where such jumps would be a pronounced issue without our proposal.

\section{Model selection}
\label{section:model_selection}
A good choice of tuning parameters is essential for accurate estimation results. The parameter $\lambda_0$ controls the form of the gain curve. When chosen too small, the glasso tends to overfit and the resulting gain curve has an inverse U shape independently of the underlying change points. When chosen too big, the glasso underfits, resulting in an almost constant gain curve. 
Even though we adjust the penalization depending on the segment length ($\lambda_{(u,v]} = \sqrt{(v-u) / n}\lambda_0$ as discussed in Section~\ref{section:complete_case}), one global $\lambda_0$ might not be able to approximate the shape (piecewise convex structure) of the population version of the gain curve simultaneously in all possible segments encountered during BS or OBS. If the sparsity pattern of the underlying graphical model in the segments differs strongly, selecting a new $\lambda_0$ in each splitting step of BS or OBS is advocated to obtain good results. We defer the details on how to locally choose the tuning parameter $\lambda_0$ (which we recommend using in practice) to the end of this section.

The parameters $\gamma$ (the penalty for the number of segments) and $\delta$ (the minimum relative segment size) on the other hand control the depth of the tree structure generated by BS (or OBS) and thus how many change points are found. A sufficiently large value of $\delta$ is also necessary to achieve stability of fits in high-dimensional scenarios. Often, overly small segments are uninterpretable and thus uninteresting for practitioners, such that $\delta$ can be set to some predetermined value (typically around 0.1).
Thus, $\gamma$ is the key parameter to be chosen in order to avoid under- or oversegmentation.
\citet{LeonBuhl} proposed to choose values for $\lambda_0$ and $\gamma$ via 2-fold cross validation, where the test data is taken from an equispaced grid across the entire sample (in this case every second observation). Since change points for $\gamma$ can be regained from BS trees grown with smaller $\gamma' < \gamma $, it is only necessary to do one BS fit (with $\gamma = 0$) per fold and $\lambda_0$.

This approach did not yield satisfactory results in our settings even when only a small amount of values were missing. Often the value chosen for $\gamma$ would correspond to the correct segmentation of the test data, but would underfit on the whole data. This could be explained by the fact that while our covariance estimation methods approximately preserve the structure of the gain curve, they do not reliably preserve its magnitude and thus $\gamma$ might be incomparable between folds and different segments.

In order to eliminate the previously mentioned issues, we propose an alternative method both as a stopping criterion for splits as well as for choosing $\lambda_0$ in practice. This approach performed empirically much better than the one that was used by \citet{LeonBuhl} and has the strong computational advantage of only requiring one single BS or OBS fit.
For each investigated segment $(u,v]$ we first apply 10-fold cross validation, taking the test data from an equispaced grid in $\uv$ (every 10-th observation in this case) to obtain an optimal value $\hat{\lambda_0}(\uv)$ corresponding to the minimal attained cross validated loss, which we denote by $l_{\uv}(\hat{\lambda_0}(\uv))$. 
The loss that is minimized is the negative log-likelihood of the test data given the mean and the estimated precision matrix of the train data.
We then use $\hat{\lambda_0}(\uv)$ for the evaluation of the gain curve for that segment. In the standard setting, one would then check if $G_n^{(u,v]}(\hat\alpha_{(u, v]}) > \gamma$ to decide whether to split further at the found point $\hat\alpha_{(u, v]}$ or not. 
Instead, we compare the cross-validated minimal loss on the full segment $\uv$ to the sum of cross-validated minimal losses of the subsegments $(u, \hat\alpha_{(u,v]}]$ and $(\hat\alpha_{(u,v]}, v]$. We keep the split 
if there is a positive improvement, i.e. if
\begin{align*}
\begin{split}
l_{\uv}(\hat{\lambda_0}(\uv)) &
- 
l_{(u,\hat\alpha_{(u,v]}]}(\hat{\lambda_0}((u, \hat\alpha_{(u, v]}]))  \\ 
&
- l_{(\hat\alpha_{(u,v]},v]}(\hat{\lambda_0}((\hat\alpha_{(u, v]},v])) > 0.
\end{split}
\end{align*}

Hence, in the decision rule for keeping a candidate, $\gamma$ is essentially removed and the decision to keep or discard a segment is just based on cross validated (out of sample) losses (in particular, whether they indicate any improvement). Note that our approach is not a proper cross-validation technique as the above described improvement at a candidate split $\hat\alpha_{(u,v]}$ is evaluated on the same data as was used to find the split point. Hence, this procedure might be optimistic regarding the improvements and thus slightly biased towards finding too many change points. This would not be a big problem in practice, as finding too many change points is preferred to finding too few. In Section \ref{section:simulations} we investigate how our stopping criterion performs on simulated data both with and without change points. Contrary to expectations it does not tend to oversegment in scenarios without change points.

\section{Simulations}
\label{section:simulations}
It seems to be hard to provide theoretical guarantees in the setting of change point detection in GGMs with missing values. We think that the only semi-realistic case to provide theory is the one with values missing completely at random, which seems to be far from realistic for most applications (see examples in Section \ref{section:applications}).
Moreover, keeping our tuning parameter $\lambda_0$ fixed across all possible segments $(u,v]$ would be one of our technical assumptions, which however is clearly suboptimal for real and simulated data (as discussed in Section \ref{section:model_selection}).
We thus focus on the practical performance of our methods, which we demonstrate with the following simulations.

\subsection{Setup}
We first discuss how we generated data, how we deleted values and we introduce the performance measure used to evaluate the results.

\paragraph{Generating covariance matrices}
In our simulations we consider two methods to draw precision matrices for the segments.
The first one is the random graph model \citep{randomgraph}, which was used to simulate high-dimensional graphs \citep[e.g. by][]{Kolar_Ising}. Here, the graph is generated by connecting nodes randomly with some probability $q  > 0$, which we set to $\frac{5}{p}$ in the following to ensure sufficient sparsity. We create the corresponding precision matrix by assigning a constant value (taken here as $0.3$) to the entries corresponding to the chosen edges and then adding the absolute value of the smallest eigenvalue of the resulting matrix plus some increment (here $0.1$) to the diagonal. This is necessary to ensure positive definiteness of the constructed precision matrix. 

We used chain networks as a second model \citep[see e.g.~Example 4.1 in][]{fan2009_chain_network}. Here we set $\Sigma_{ij} := \exp{(-a|s_i-s_j|)}$, where $a > 0$ , $s_1 < \ldots  < s_p$ and $s_i - s_{i - 1} \sim \Unif(0.5,1)$ for $i = 2,\dotsc,p$. In the simulations we set $a = 1/2$ and additionally draw $s_1\sim \Unif(0.5, 1)$. The inverse of the resulting matrix is tridiagonal. To generate precision matrices with differing sparsity patterns for different segments, we draw some permutation $\pi$ of $1,\dotsc, p$ and set $\Sigma_{ij} = \exp(-a|s_{\pi(i)} - s_{\pi(j)}|)$. This breaks the tridiagonal structure but keeps the sparsity.

\paragraph{Types of missingness}
We consider two ways to delete values for the simulations. The first one is missing completely at random (MCAR), where a given percentage of values is discarded uniformly at random. The second one is inspired by the missingness structure of environmental monitoring data (see bottom of Figure \ref{fig:groundwater_final}). Here a failure of a sensor leads to missing values over several consecutive observations, while replacement of them at multiple sampling locations might occur at the same time. Moreover, it is common that simultaneously several sites are newly installed or abandoned based on the available budget. This leads to blocks of observations missing, ranging over multiple variables as well as observations. In order to generate a similar blockwise missingness structure, we repeatedly select $k \sim  \Poi(\frac{p}{20})$ variables uniformly at random and delete for all of them a segment of length $l \sim  \Exp(\frac{n}{8})$ with the midpoint chosen uniformly between $1$ and $n$. We repeat this procedure until the preset percentage of missing values is reached. An example with 30\% missing values for $n = 200$ and $p = 100$ is shown at the bottom of Figure \ref{fig:curve_smoothing}.

\paragraph{Performance measures}
We use the adjusted Rand Index \citep{adjRand_Hubert}, a common measure to compare clusterings, to measure performance regarding change point detection in our simulation study (see Table \ref{table:simulation_results}). For two partitions of $n$ observations, the Rand Index  \citep{Rand} is the number of agreements (pairs of observations that are either in the same subset for both partitions or are in different subsets for both partitions) divided by the total number of pairs~$n\choose{2}$.
The adjusted Rand Index is the difference between the Rand Index and its expectation when choosing partitions randomly, normalized by the difference between the maximum possible Rand Index and its expectation. The adjusted Rand Index is thus bounded by one and is expected to be zero when partitions are chosen randomly. We illustrate the estimation uncertainty of found change points corresponding to some adjusted Rand measures (taking true and estimated segments as the two partitions) via histograms in Figure~\ref{fig:histograms}.

An interesting point includes also the accuracy in recovering the precision matrices (or the underlying graphical models). The reasons why we only focus on the recovery of change points and do not report performance measures related to graph recovery are the following.
We tuned our algorithm specifically for good detection of the change points and not necessarily for good estimation of the precision matrices (see the discussion in Section~\ref{sec:adapting_to_missing_data}), because good recovery of the graph structure is impossible without precisely knowing the change point locations, partly due to the precision matrix of the mixture distribution not being sparse in many cases. 
Of course, once change points are accurately localized, one can use the most favoured imputation method for homogeneous data (e.g.~Miss-Glasso by \cite{missglasso}, the proposals of \cite{loh2012structure} or \cite{Kolar_missing}, or even procedures that are not focused specifically on precision matrices such as the missForest of \citealp{missforest}) to subsequently estimate the precision matrices or the underlying graphical models. The quality of graph recovery then mainly depends on the performance of the chosen imputation method.

\paragraph{Setup of the main simulation study}
We will illustrate the behavior of our methods on settings with $n = 500$ observations of dimension $p = 100$ with three change points with segments of sizes $70,120,120$ and $190$. We randomly permute the order of the segments to avoid systematic effects. Note that in the smallest segment the number of observations is smaller than the number of variables, resulting in a truly high-dimensional setting when splitting. In each simulation, we generate a precision matrix (either random or chain network) for each segment and then draw observations independently from the corresponding centered multivariate normal distribution.
The parameter $\delta$ is held fixed at $0.1$ and we vary the proportion of missing data in steps of $10\%$ between $10\%$ and $50\%$. 
Note that in this setup in expectation there is less than one complete observation available per segment when deleting only $10\%$ of the values completely at random. Therefore, discarding incomplete observations is clearly not a viable option and some kind of imputation method is necessary.

\subsection{Results}
\label{section:simulation_results}
\begin{figure}
    \centering
    \includegraphics[width = 230pt]{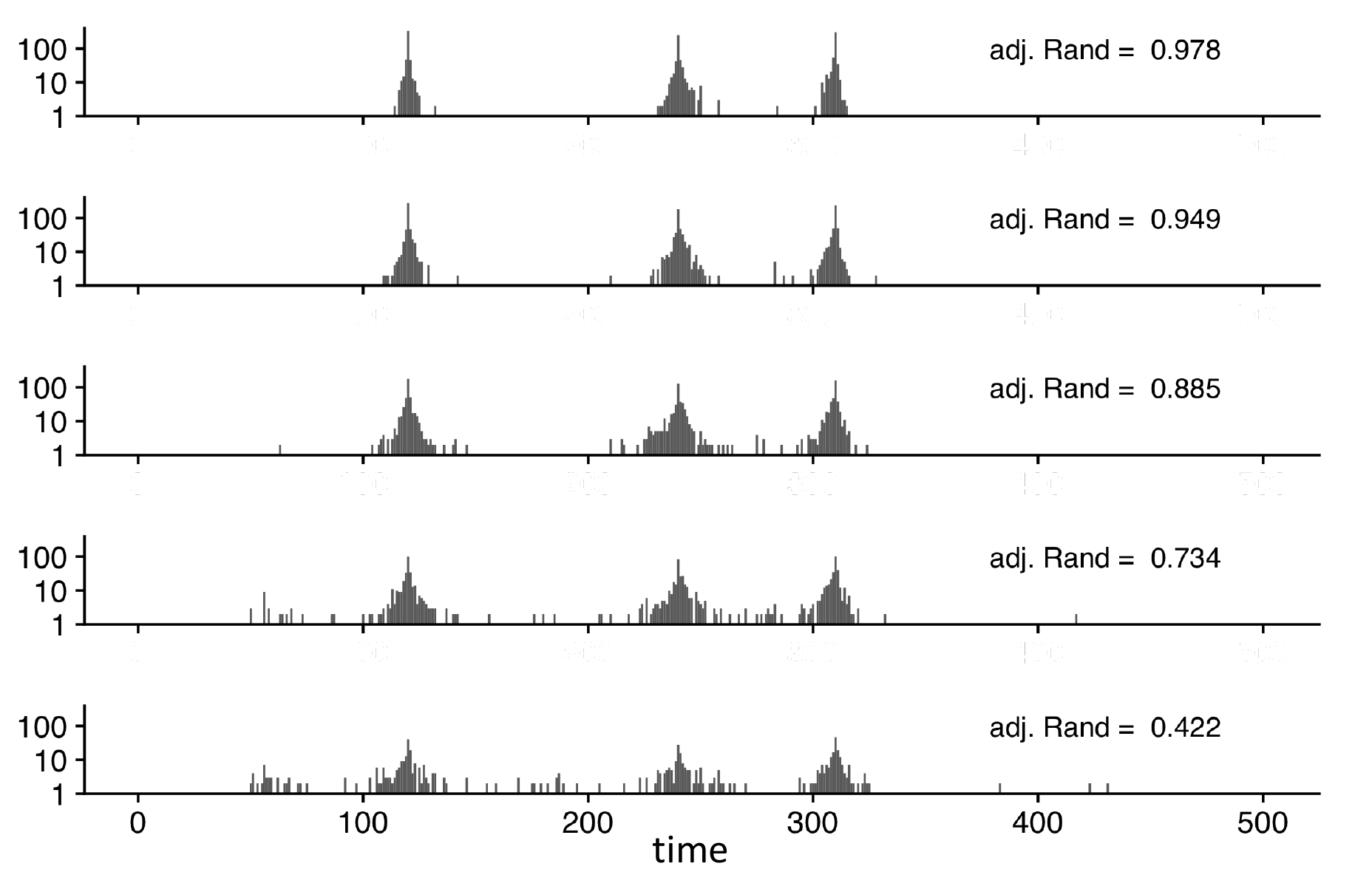}
    \caption{Cumulative estimated change points from each 500 simulations with true underlying change points at 120, 240 and 310 and random networks as the in segment precision matrices. We deleted 10\% (top) to 50\% (bottom) of the values blockwise and estimated the change points using OBS and the LW method. Also displayed are the average adjusted Rand Indices for each set of simulations.}
    \label{fig:histograms}
\end{figure}

We analyze the estimation performance (using the model selection approach of Section \ref{section:model_selection}) for our three methods (average, pairwise and LW, see Section \ref{section:imputation_methods}) both using BS and OBS. We ran 100 simulations for each setting. The corresponding mean values of adjusted Rand Indices are displayed in Table \ref{table:simulation_results} along with their standard deviations in parenthesis. To aid the interpretability, we present the adjusted Rand Indices of a selection of estimation results together with the true change points in Table \ref{table:Rand_example}. Note that finding the correct number of change points with an accuracy of around two observations each leads to an adjusted Rand Index of around 0.95. Finding all true change points plus a false positive one leads to an adjusted Rand Index of around 0.8, similar to finding only two of the three change points.
Additionally we show histograms of all the change points found over 500 simulations in Figure~\ref{fig:histograms}, for which we needed to consider a fixed true change points scenario to be meaningful. Note that the y-axis is on a log-scale. 

\begin{table*}
\centering
\caption{Adjusted Rand Indices for scenarios with $n = 500$, $p = 100$ and three change points}
\vspace{0.08cm}
\begin{tabular}{lllllllll}
\label{table:simulation_results}
 &  &  &  & \multicolumn{5}{c}{\bf PERCENTAGE OF MISSING VALUES} 
 \vspace{0.08cm} \\
\multicolumn{1}{c}{\bf NW} & \multicolumn{1}{c}{\bf MISS} & \multicolumn{2}{c}{\bf METHOD} &  \multicolumn{1}{c}{\bf 10\%} & \multicolumn{1}{c}{\bf 20\%} & \multicolumn{1}{c}{\bf 30\%} & \multicolumn{1}{c}{\bf 40\%}  & \multicolumn{1}{c}{\bf 50\%}\\ \hline
CN & MCAR & LW & BS & 0.999 (0.003) & 0.996 (0.006) & 0.993 (0.010) & 0.960 (0.060) & 0.430 (0.314) \\ 
  CN & MCAR & LW & OBS & 0.998 (0.004) & 0.996 (0.006) & 0.987 (0.025) & 0.949 (0.068) & 0.466 (0.301) \\ 
  CN & MCAR & av & BS & 1.000 (0.001) & 0.998 (0.004) & 0.952 (0.072) & 0.244 (0.256) & 0.000 (0.000) \\ 
  CN & MCAR & av & OBS & 0.999 (0.002) & 0.998 (0.005) & 0.957 (0.069) & 0.226 (0.252) & 0.000 (0.000) \\ 
  CN & MCAR & pair & BS & 0.999 (0.003) & 0.997 (0.006) & 0.993 (0.010) & 0.939 (0.118) & 0.372 (0.297) \\ 
  CN & MCAR & pair & OBS & 0.998 (0.004) & 0.996 (0.006) & 0.986 (0.026) & 0.938 (0.074) & 0.384 (0.290) \\ 
  CN & block & LW & BS & 0.999 (0.004) & 0.994 (0.012) & 0.985 (0.020) & 0.953 (0.051) & 0.813 (0.208) \\ 
  CN & block & LW & OBS & 0.998 (0.005) & 0.991 (0.019) & 0.977 (0.037) & 0.923 (0.085) & 0.727 (0.257) \\ 
  CN & block & av & BS & 0.957 (0.086) & 0.811 (0.214) & 0.367 (0.340) & 0.242 (0.267) & 0.280 (0.245) \\ 
  CN & block & av & OBS & 0.953 (0.123) & 0.765 (0.243) & 0.294 (0.326) & 0.197 (0.268) & 0.248 (0.253) \\ 
  CN & block & pair & BS & 0.998 (0.006) & 0.990 (0.017) & 0.981 (0.025) & 0.941 (0.069) & 0.813 (0.170) \\ 
  CN & block & pair & OBS & 0.997 (0.007) & 0.990 (0.013) & 0.958 (0.039) & 0.923 (0.076) & 0.732 (0.240) \\ 
  RN & MCAR & LW & BS & 0.991 (0.012) & 0.973 (0.051) & 0.943 (0.083) & 0.828 (0.116) & 0.510 (0.228) \\ 
  RN & MCAR & LW & OBS & 0.986 (0.025) & 0.971 (0.046) & 0.932 (0.090) & 0.791 (0.146) & 0.452 (0.270) \\ 
  RN & MCAR & av & BS & 0.981 (0.041) & 0.878 (0.102) & 0.578 (0.221) & 0.076 (0.184) & 0.005 (0.050) \\ 
  RN & MCAR & av & OBS & 0.983 (0.028) & 0.858 (0.120) & 0.499 (0.275) & 0.063 (0.168) & 0.011 (0.075) \\ 
  RN & MCAR & pair & BS & 0.988 (0.014) & 0.974 (0.050) & 0.931 (0.086) & 0.779 (0.141) & 0.438 (0.253) \\ 
  RN & MCAR & pair & OBS & 0.986 (0.025) & 0.970 (0.050) & 0.917 (0.094) & 0.766 (0.156) & 0.373 (0.266) \\ 
  RN & block & LW & BS & 0.990 (0.021) & 0.974 (0.038) & 0.927 (0.103) & 0.791 (0.208) & 0.464 (0.328) \\ 
  RN & block & LW & OBS & 0.979 (0.041) & 0.949 (0.073) & 0.904 (0.137) & 0.741 (0.225) & 0.386 (0.342) \\ 
  RN & block & av & BS & 0.752 (0.210) & 0.336 (0.311) & 0.173 (0.250) & 0.198 (0.238) & 0.276 (0.226) \\ 
  RN & block & av & OBS & 0.711 (0.243) & 0.234 (0.287) & 0.119 (0.207) & 0.136 (0.224) & 0.227 (0.244) \\ 
  RN & block & pair & BS & 0.986 (0.027) & 0.967 (0.049) & 0.923 (0.100) & 0.759 (0.221) & 0.573 (0.313) \\ 
  RN & block & pair & OBS & 0.980 (0.037) & 0.949 (0.079) & 0.863 (0.181) & 0.700 (0.234) & 0.483 (0.339) \\ 
\end{tabular}
\end{table*}

\begin{table}
\caption{Examples for different adjusted Rand Indices}
\label{table:Rand_example}
\begin{center}
\begin{tabular}{l l l}
\multicolumn{1}{c}{\bf TRUE CPTS} & \multicolumn{1}{c}{\bf FOUND CPTS} & \multicolumn{1}{c}{\bf ADJ. RAND} \\
\hline 
120, 190, 310          & 120, 188, 310 & 0.992 \\
70, 190, 310           &         66, 191, 310 & 0.980\\
120, 240, 430    &               118, 239, 423 & 0.9
49 \\  
190, 260, 380 &               93, 190, 260, 380 &  0.804\\
120, 240, 310 &                      119, 243 & 0.757\\
120, 240, 310           &                297& 0.506\\
70, 190, 380 &                         380 & 0.363
\end{tabular}
\end{center}
\end{table}

Before analyzing the obtained results, we first report the performance of a naive approach (for some of the setups from Table~\ref{table:simulation_results}). For this, we impute mean values columnwise (i.e.~for each variable separately) on the whole data set and then estimate change points using the imputed data (with our proposed methodology and implementation for the case when there are no missing values). Note that in our simulations (i.e., the setups of Table~\ref{table:simulation_results}), there are no changes in the underlying  mean values. Hence, one could expect (at least in the scenario of values missing completely at random) this naive approach to perform similar to the average imputation method. For chain networks with values missing completely at random, the naive approach performs similar to our three proposed imputation methods for up to $30\%$ missing values. For $40-50\%$ missing values, the naive approach performs better than the average imputation method, but clearly worse than the best performing pairwise and LW methods (by roughly $0.15$ in terms of adjusted Rand Index). For chain networks with blockwise missing values (and Optimistic Binary Segmentation), the naive approach obtained $0.61, 0.49, 0.40, 0.38$ and $0.34$ for the adjusted Rand Indices for the scenarios with $10$ to $50\%$ missing values, respectively. For low percentage of missing values this is clearly worse than all three of our imputation proposals (by roughly $0.4$ in terms of adjusted Rand Index) and for high percentage of missing values it is roughly as good as the average imputation method, and clearly worse than the best performing pairwise and LW methods (by roughly $0.4$ in terms of adjusted Rand Index). The bad performance of the naive approaches can be explained as follows. When values are not missing completely at random, but in some kind of blockwise structure, the naive approach tends to (falsely) find change points corresponding to the missingness structure rather than true changes in the underlying signal. Similar issues with the naive approach could happen when there are additionally also shifts in the mean of the signal. Hence, integrating imputation and change point detection into a unified framework is essential and leads to substantial gains, especially in very challenging scenarios (with values not missing completely at random and/or high percentage of missing values).

With regards to the main simulation results comparing our three proposals, all three methods perform similarly if at most 30\%  of values for the chain network setup or 20\% of values for the random network setup are deleted completely at random. For more challenging MCAR setups the average imputation method fails to estimate the change points reliably, whereas the LW and pairwise methods still perform reasonably well if up to 30\% -- 40\% of values are missing. The performance of all methods tends to be worse if the values are deleted blockwise, as expected. Here, the average imputation method clearly underperforms compared to the other two methods even when only 10\% of the values are deleted. The pairwise and LW methods perform very well for up to 30\% -- 40\% of missing values. The LW method seems to perform somewhat better than the pairwise method. The scenarios with 50\% missing values are all very challenging and the ranking of the methods may be slightly different compared to the easier scenarios.

\begin{figure}
    \centering
    \includegraphics[width = 250pt]{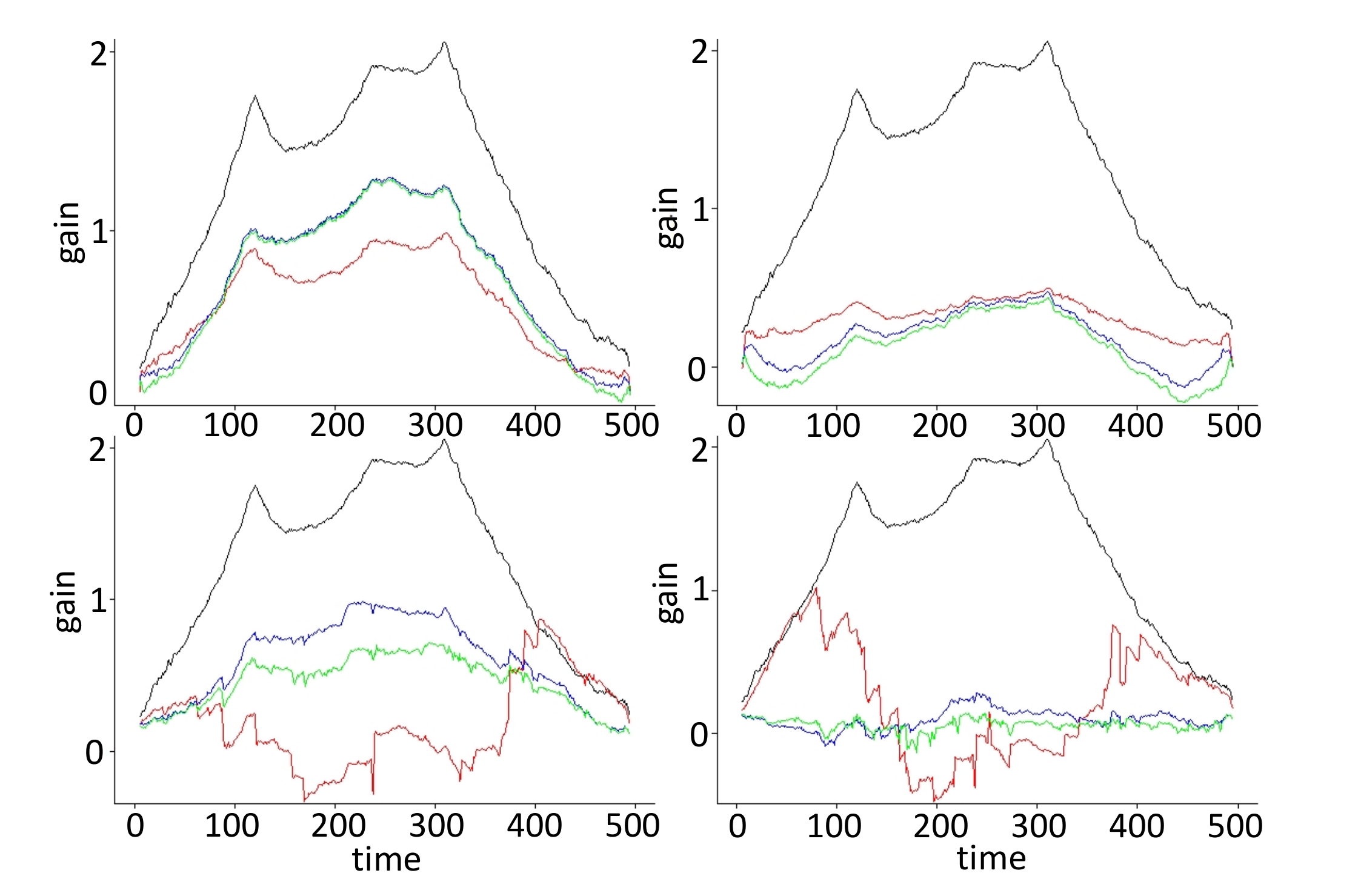}
    \caption{Recovered gain curves in random network settings with 30\% MCAR (top left), 30\% blockwise (bottom left), 50\% MCAR (top right) and 50\% blockwise (bottom right) missing values using the average (red), pairwise (green) and LW (blue) methods, compared to the case without missing values (black).}
    \label{fig:compare_imputations}
\end{figure}

Figure~\ref{fig:compare_imputations} provides representative examples comparing the gain curves recovered using our three methods. If 30\% of the values are deleted completely at random, the expected piecewise convex shape is well conserved for all three methods. For 30\% blockwise deleted values the shape is somewhat distorted when using the LW and pairwise methods but local optima still occur at two of the true change points. The structure (i.e.~shape) is not recoverable at all due to big jumps with the average imputation method, explaining the low mean adjusted Rand Index in this scenario. A key insight is that with the LW and pairwise  methods the piecewise convex structure is approximately conserved, allowing for good estimation of at least one of the underlying change points, as necessary for BS and OBS. Some very challenging scenarios with 50\% missing values are shown in the right panel of Figure~\ref{fig:compare_imputations}. Here even in the MCAR scenario (top right), the gain curves flatten out, making it difficult to accurately locate the change points, while in the blockwise missing scenario (bottom right) all three methods would fail. Note that the imputation method (and in general also the fraction of missing values) has an effect on the magnitude of the gain curve.

The simulation scenarios with random networks seem to be harder for our methods to estimate change points compared to chain networks. As expected, at the price of higher computational cost, BS slightly outperforms OBS, especially in hard scenarios with a high fraction of missing values where we expect the piecewise convex shape of the gain curve to be strongly distorted.
We also tested standard BS and OBS (as described in Section \ref{section:complete_case}) on scenarios where no values are missing, but the total number of observations is scaled down by 10\% -- 50\% (Table~\ref{table:simulation_results_comparison}). Not surprisingly, we see that complete data are generally more informative than the analogous data with missing values. Also here, the random network scenarios are more challenging compared to chain networks.

\begin{table*}
\centering
\caption{Adjusted Rand Indices for scenarios with $p = 100$, three change points and no missing values}
\vspace{0.08cm}
\label{table:simulation_results_comparison}
\begin{tabular}{lllllll}
 &  &  \multicolumn{5}{c}{\bf NUMBER OF OBSERVATIONS} 
 \vspace{0.08cm} \\
\multicolumn{1}{c}{\bf NW} & \multicolumn{1}{c}{\bf METHOD} &  \multicolumn{1}{c}{\bf 450} & \multicolumn{1}{c}{\bf 400} & \multicolumn{1}{c}{\bf 350} & \multicolumn{1}{c}{\bf 300}  & \multicolumn{1}{c}{\bf 250}\\ \hline
CN & BS & 1.000 (0.001) & 1.000 (0.001) & 0.999 (0.002) & 1.000 (0.002) & 0.997 (0.009) \\ 
  CN & OBS & 1.000 (0.002) & 0.999 (0.003) & 0.998 (0.006) & 0.999 (0.004) & 0.996 (0.016) \\ 
  RN & BS & 0.990 (0.019) & 0.990 (0.018) & 0.980 (0.039) & 0.947 (0.077) & 0.875 (0.095) \\ 
  RN & OBS & 0.987 (0.021) & 0.987 (0.019) & 0.969 (0.051) & 0.931 (0.083) & 0.833 (0.084) \\ 
\end{tabular}
\end{table*}

In the middle scenario of Figure~\ref{fig:histograms} with 30\% missing values we estimate a total of 1343 change points with 1079 within five and 1190 within ten observations away from a true underlying change point. The analogous numbers for the settings with 40\% missingness are 1118, 754 and 882 and for 50\% missingness are 724, 353 and 447. The decreasing performance is also reflected in the average adjusted Rand Indices. In total there would be $3\cdot500=1500$ true underlying change points. This indicates a general trend, where our algorithm seems to underfit, i.e. selecting too few change points, rather than selecting incorrect ones. 
To investigate our model selection procedure presented in Section \ref{section:model_selection}, we did a complementary study in a setting of no true underlying change points. Testing all scenarios of Table \ref{table:simulation_results} with 100 simulations each resulted in a total of $24 \cdot 5 \cdot 100 = 12000$ simulations. In the setting where $n = 500$ and $p = 100$ and without true underlying change points our estimators found a total of 734 change points, 728 of which were found with the average imputation method for data where values were deleted blockwise, four with the pairwise method and two with the LW method, both with blockwise deleted values.
Our model selection did not select any change points in any of the 6000 simulation with values deleted completely at random or any of the 4800 simulations with the LW or pairwise estimator where up to 30\% of values were deleted blockwise. In an analogous simulation study with $n = 100$ and $p = 100$ (setting $\delta = 0.2$ to obtain reasonable fits) again no change points were found in the simulations with values missing completely at random. In total 615 change points were found in the 12000 simulations. 109 were found with the average method, 285 were found with the LW method and 221 with the pairwise method. Only 3, 1 and 3 respectively were found in scenarios with up to 30\% missing values. 

Finally we indicate the computational cost of the different methods. Running on a single Intel Xenon 3.0 GHz processor core, with our current \textsf{R} implementation each BS search including model selection took around 130, 150 and 170 seconds for the average, LW and pairwise method for an average scenario of Table~\ref{table:simulation_results}. Change point estimation with OBS took around 40 seconds for each of the three methods, thus enabling massive speed-ups.

\section{Applications}
\label{section:applications}
In every real application care needs to be taken as the assumption of underlying piecewise constant GGMs might not be fully valid. In particular, the underlying model might also change smoothly over time. Such deviations from the model assumptions are usually visible in the shape of the gains curves. Other deviations from the model assumptions could be dependence across observations for example. In this case various pre-processing steps (e.g.~smoothing, differencing or various transformations) might be helpful to provide more reliable results.

\begin{figure}[h] 
    \centering
    \includegraphics[width = 230pt]{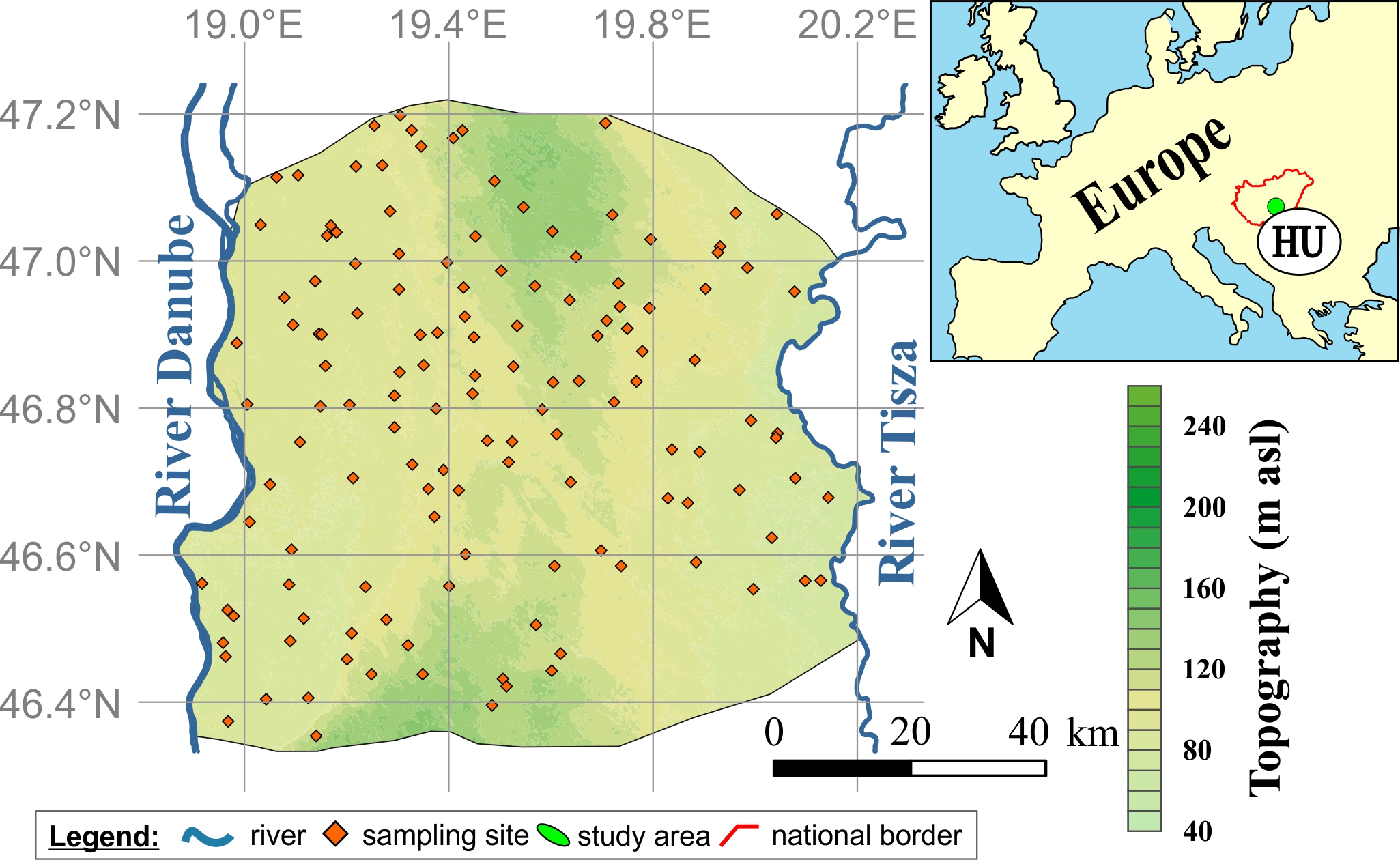}
    \caption{Location map of the shallow groundwater monitoring area within Hungary.}
    \label{fig:groundwater_area}
\end{figure}

We applied our methods on a real data example from a shallow groundwater monitoring system with naturally occurring missing values. The data set contains $n = 753$ monthly shallow groundwater level measurements at $p=136$ sampling locations between the Rivers Danube and Tisza within Hungary (Figure \ref{fig:groundwater_area}) from January 1951 to September 2013. The original measurements were seasonally adjusted for each sampling location individually. This data set has approximately 35\% naturally occurring missing values.

\begin{figure*}[ht]
\hspace*{-1cm}   
    \centering
    \includegraphics[width = 450pt]{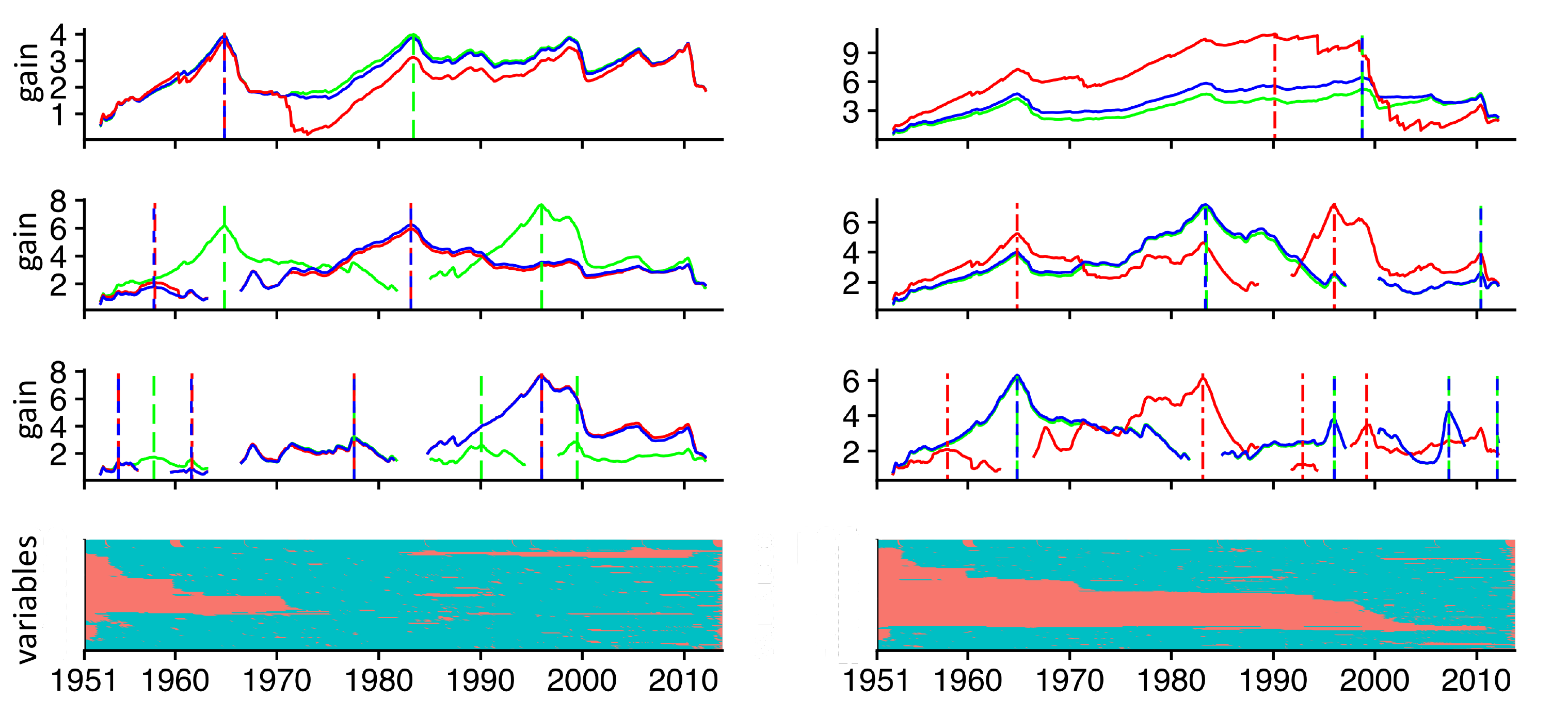}
    \caption{Gain curves of the first three BS iterations with the average imputation (red), LW (blue) and pairwise (green) methods applied to the full groundwater data (right) and to the subset only including sampling locations with less than 50\% missing values (left). Splits found in each step are marked with a vertical line of the corresponding color and the gain curves of the subsegments are shown in the plot below. The missingness structure of the data sets is displayed in the bottom row with red parts corresponding to missing values.
    }
    \label{fig:groundwater_final}
\end{figure*}

Due to the nature of the monitoring system there are a lot of missing values with a blockwise missingness structure that is strongly inhomogeneous in time, see bottom right of Figure \ref{fig:groundwater_final}. The reasons for missing observations in similar monitoring systems could be the following. The failure of a sensor can lead to missing values over several consecutive observations, while replacement of them at multiple sampling locations might occur at the same time. More importantly, it is common that several sites are simultaneously newly installed or abandoned based on the available budget or even due to changing standards towards monitoring systems. These typically lead to blocks of missing observations, ranging over multiple sites (variables) as well as observations. In addition, individual randomly missing entries can occur as well.

Out of the 136 sampling sites only 93 had non-missing values for more than $n / 2 \approx 377$ months, with some having values missing for three quarters of the months. The reduced set of 93 sites only has approximately 16\% missing values. We first applied BS with our three proposed methods to this subset (left of Figure \ref{fig:groundwater_final}). We used cross-validated $\lambda_0$ for each segment as described in Section \ref{section:model_selection} and a small $\delta = 0.025$ for better visualisation of the gain curves at the boundaries. Note that plotting the whole gain curves requires a full search, for which BS (including the proposed model selection) took roughly 30 minutes. In many other scenarios (e.g. financial data) both the number of observations as well as the number of variables can be larger, quickly bringing the full grid search beyond what is computationally feasible. In such cases the faster OBS still remains a viable option.

While the order of the splits are different across the three methods, the finally found change points lie very close to each other. The three change points which all methods agree on for both the full and reduced data set with the highest cross-validated improvements lie around November 1964, April 1983 and January 1996 (with up to only two observations difference). Besides these three, several smaller ones are found. The major change points have a clear hydrogeological interpretation. For example, around 1983, there was a drop in groundwater levels unequally impacting areas within the monitoring system. This presumably shows up jointly as a mean and covariance shift in our analysis. Note that our log-likelihood based approach will also detect shifts in mean.

Interestingly, applying BS to the very challenging full data set yields very similar results (right of Figure \ref{fig:groundwater_final}). This would not have been possible without the techniques preventing jumps in the gain curves discussed in Section \ref{section:curve_smoothing} due to the big  blocks of missing values. Even with our technique, the gain curve for the first split flattens out, showing the challenges in such complicated scenarios. Also note the jumps in the gain curve of the average imputation method around the boundaries of big missing blocks, similar to Figure~\ref{fig:compare_imputations}.

\section{Conclusions}
Our estimation methods enable practitioners to search for change points in settings where this was impossible before. The Loh-Wainwright based imputation method presented in Section \ref{section:imputation_methods} has a stable performance even in challenging high-dimensional scenarios with lots of missing values. Since its computational cost is not significantly higher than the other methods considered, we generally recommend its usage. The choice of BS or OBS should be based on computational resources. BS results in slightly better results and the possibility of better visualization by drawing the full gain curves, however at a considerably higher computational cost than OBS. We emphasize that technical adjustments regarding the evaluation of the gains introduced in Section~\ref{section:curve_smoothing} as well as the model selection procedure of Section~\ref{section:model_selection} lie at the core of our methodology, enabling good performances both on simulated and real data with missing values.

\subsubsection*{Acknowledgements}
We thank %an Associate Editor and two 
anonymous reviewers for constructive comments. Furthermore, we thank Tam\'as Garamhegyi, J\'ozsef Kov\'acs (Department of Geology, E\"otv\"os Lor\'and University, Budapest, Hungary) and J\'ozsef Szalai (General Directorate of Water Management, Budapest, Hungary) for the groundwater data set and for providing hydrogeological interpretation of the found change points. Solt Kov\'acs and Peter B\"uhlmann have received funding from the European Research Council (ERC) under the European Union's Horizon 2020 research and innovation programme (Grant agreement No.~786461 CausalStats - ERC-2017-ADG).

\bigskip
\begin{center}
{\large\bf Supplementary materials}
\end{center}
%\begin{description}
%\item[\textsf{R}-package \textbf{hdcd}:] 
The \textbf{\textsf{R}-package} \textbf{hdcd} for \textbf{h}igh-\textbf{d}imensional \textbf{c}hange point \textbf{d}etection available at \url{https://github.com/mlondschien/hdcd} implements our methods to find change points in GGMs with possibly missing values. README files explain the usage via some examples and describe how to reproduce the above mentioned simulation results and figures. 
Additionally, the \textbf{hdcd} \textsf{R}-package also implements some ongoing work on multivariate nonparametric change point detection (see \citealp{RF_change_point}) and is planned to be uploaded also to CRAN later on.
%\end{description}

%added 2020a, 2020b and 2020c into bibliography file directly to make articles distinguishable
\bibliography{bib_arxiv}

\end{document}